%% file: main.tex
\documentclass[letterpaper]{article} 
\usepackage[submission]{aaai24}  
\usepackage{times}  
\usepackage{helvet}  
\usepackage{courier}  
\usepackage[hyphens]{url}  
\usepackage{graphicx} 
\usepackage{natbib}  
\usepackage{caption} 
\usepackage{algorithm}
\usepackage{algorithmic}
\usepackage{multirow}
\usepackage{booktabs}
\usepackage{newfloat}
\usepackage{listings}
\DeclareCaptionStyle{ruled}{labelfont=normalfont,labelsep=colon,strut=off} 
\floatstyle{ruled}
\newfloat{listing}{tb}{lst}{}
\floatname{listing}{Listing}
\title{CLeaRForecast: Contrastive Learning of High-Purity Representations for Time Series Forecasting}

\author{
    Jiaxin Gao\textsuperscript{\rm 1} \textsuperscript{\rm 2},
    Yuxiao Hu\textsuperscript{\rm 2},
    Qinglong Cao\textsuperscript{\rm 1} \textsuperscript{\rm 2},
    Siqi Dai\textsuperscript{\rm 2},
    Yuntian Chen\textsuperscript{\rm 2}\equalcontrib
}
\affiliations{
    \textsuperscript{\rm 1} Shanghai Jiao Tong University
\textsuperscript{\rm 2 }Ningbo Institute of Digital Twin, Eastern Institute of Technology, Ningbo

}

\usepackage{bibentry}

\begin{document}

\maketitle

\begin{abstract}
Time series forecasting (TSF) holds significant importance in modern society, spanning numerous domains. Previous representation learning-based TSF algorithms typically embrace a contrastive learning paradigm featuring segregated trend-periodicity representations. Yet, these methodologies disregard the inherent high-impact noise embedded within time series data, resulting in representation inaccuracies and seriously demoting the forecasting performance. To address this issue, we propose CLeaRForecast, a novel contrastive learning framework to learn high-purity time series representations with proposed sample, feature, and architecture purifying methods. More specifically, to avoid more noise adding caused by the transformations of original samples (series), transformations are respectively applied for trendy and periodic parts to provide better positive samples with obviously less noise. Moreover, we introduce a channel independent training manner to mitigate noise originating from unrelated variables in the multivariate series. By employing a streamlined deep-learning backbone and a comprehensive global contrastive loss function, we prevent noise introduction due to redundant or uneven learning of periodicity and trend. Experimental results show the superior performance of CLeaRForecast in various downstream TSF tasks. 
\end{abstract}

\section{Introduction}

Time series forecasting (TSF) aids in uncovering hidden patterns and trends inherent in time series data, enabling accurate predictions for the future. It serves as a useful tool in various domains such as financial market analysis \cite{lopez2023can}, energy management \cite{gao2023adaptive}, transportation planning \cite{fang2022attention}, and to name a few. Recent developments in deep learning models have showcased notable advancements in TSF \cite{Zeng2022AreTE} \cite{haoyietal-informer-2021} \cite{gao2023client}. Among these advancements, self-supervised learning methods, especially contrastive learning methods \cite{tonekaboni2021unsupervised} \cite{yue2021ts2vec} \cite{woo2022cost}, have emerged as one of the promising methods in the realm of TSF tasks. Contrastive learning methods learn representations by mapping similar sub-series (i.e., positive pairs) to similar representations while pushing dissimilar sub-series (i.e., negative pairs) apart, and the representations of the series can be applied to multiple downstream tasks, such as classification, forecasting, and anomaly detection.

\begin{figure}[!t]
\centering
\includegraphics[width=1.05\columnwidth]{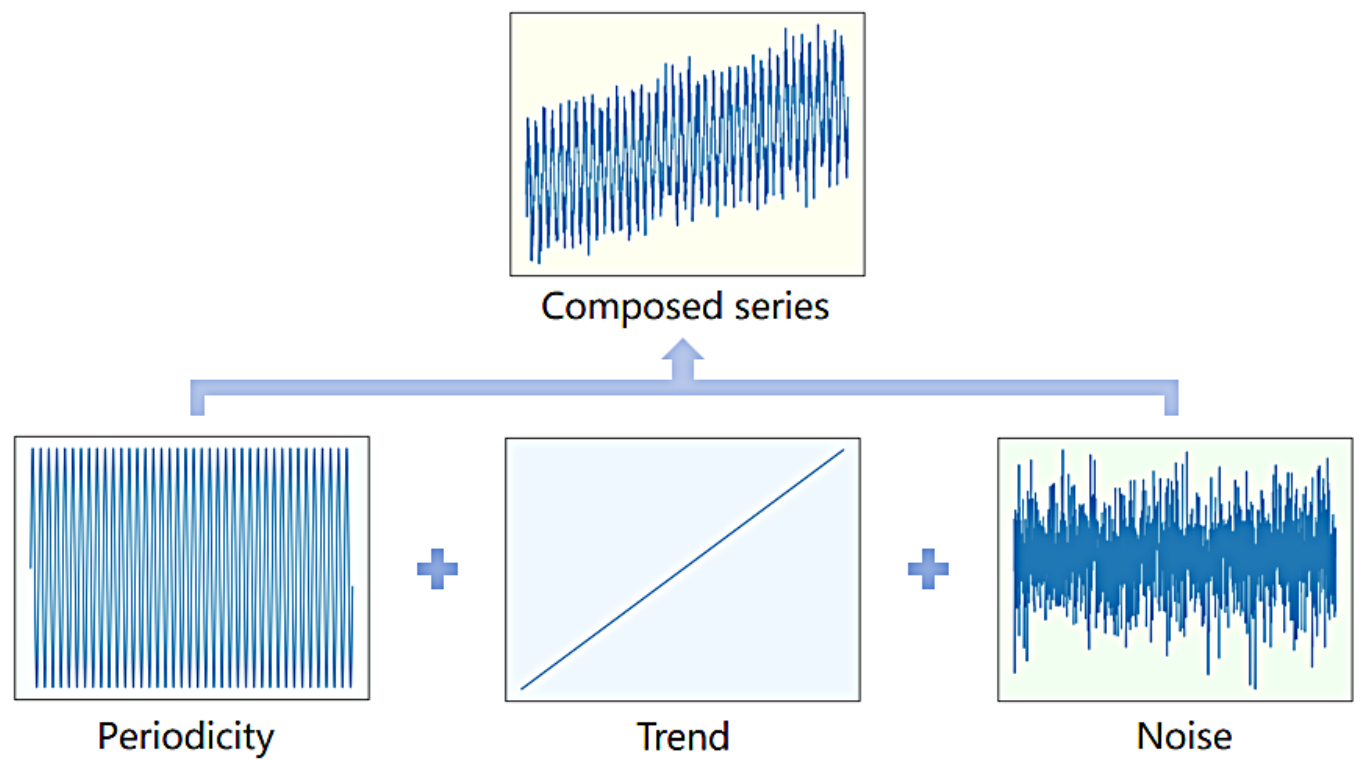} 
\caption{Time series decomposition. A time series can be decomposed into a trendy part, a periodic part, and a noise part.}
\label{decompose}
\end{figure}

 Time series can be regarded as entangled data structures, which are assumed to consist of three parts: a trendy part, a periodic (seasonal) part, and a noise part \cite{bandara2021mstl}, as shown in Fig. \ref{decompose}. Given the unpredictable nature of noise, the attainment of optimal forecasting performance relies on the ability to uncover and understand both periodic patterns and underlying trends within the time series. Therefore, learning high-purity representations of trendy and periodic patterns is crucial for time series contrastive learning methods. 

 Previous time series contrastive learning methods, like CoST \cite{woo2022cost}, have mostly overlooked the purifying of time series, including suboptimal selection of positive sample pairs, ineffective training manner with channel mix manner, and inadequacies in the design of the backbone structure and contrastive loss functions. More specifically, in generating positive samples, they only apply simple transformations of scaling, shifting and jittering to the original series, aiming to enhance the diversity of the positive samples. However, these transformations without consideration of noise would also boost the diversity of the noise and correspondingly contaminate the representations. Moreover, the existing contrastive learning frameworks predominantly utilize the channel mix training manner, where they simultaneously process multiple variables from the multivariate series in order to acquire global representations. However, it is worth noting that certain variables within a multivariate series may not exhibit significant correlations or causal relationships. Consequently, the inclusion of unrelated variables during the analysis becomes a source of disruptive noise, leading to inaccurate dependency relationships within the channel mix manner.
Additionally, previous contrastive learning backbones commonly use dilated convolutions followed by trend/periodicity encoders to learn trend/periodicity features. However, dilated convolutions, due to their long receptive field, already capture trend/periodicity features to a certain extent. The redundant learning of these features would introduce noise and impede the extraction of trend/periodicity information. Furthermore, previous contrastive learning methods usually employ multiple contrastive losses to optimize the representation in diverse perspectives of the whole representation. However, multiple contrastive losses would make the optimization more challenging. It could easily cause the representations to prioritize one perspective over another and generate exaggerated or underestimated trendy/periodic representations, which could not meet the global representation utilization requirement of the following TSF tasks.
 
To tackle these problems, we propose CLeaRForecast, a contrastive learning framework to learn high-purity representations for TSF. In this framework, following the previous trend-periodicity segregated-analyzed pipeline \cite{woo2022cost}, a range of purifying methods from sample, feature, and architecture perspectives are proposed to obtain the high-purity representation of time series. First, transformations are respectively applied to trendy and periodic parts, which eliminate the effect of the noise part and provide better positive samples with obviously less noise. Moreover, to decouple the unrelated variables in the times series, a fresh channel independent training manner is proposed to obtain the disturbing-noise-eliminating features. Furthermore, we eliminate the dilated convolution feature extractor to avoid redundant extraction of trend/periodicity features, which provides a streamlined backbone with less noise. To provide more comprehensive global representations for the downstream TSF tasks, the global contrastive loss function is proposed to promote better optimization of the model in a global manner. These purifying optimizations effectively prevent the introduction of additional noise in the representation of time series. The efficient time series representations learned by the CLeaRForecast have significantly improved performance when applied to downstream forecasting tasks.

In summary, the contribution of the proposed CLeaRForecast is summarized as follows:

\begin{itemize}

\item To the best of our knowledge, CLeaRForecast is the first contrastive learning framework to learn high-purity representations of time series data, which efficiently boosts the performance of the downstream TSF tasks.

\item To obtain high-purity time series representations, we proposed fresh purifying methods from sample, feature, and architecture aspects, including a more efficient positive sample generation strategy, channel independent training manner, and a more streamlined backbone structure with a global contrastive loss.

\item CLeaRForecast is evaluated on multiple real-world datasets. The experimental results demonstrate that our method achieves state-of-the-art performance for representation learning-based TSF methods and extensive ablation studies are conducted to testify its characteristics.



\end{itemize}

\section{Related Work}
\subsection{Time Series Forecasting Models}
Traditional time series forecasting (TSF) models such as ARMA and ARIMA \cite{box2015time} are based on statistics, and they assume linear relationships between past and present observations to identify patterns in the time series. However, deep learning models have gained popularity because they have better expressive ability and can effectively utilize the available data in the field of time series, resulting in improved performance compared to traditional statistical models. Temporal convolution network (TCN) is an important branch of deep learning models, and some TCN based models \cite{Oord2016WaveNetAG} \cite{Sen2019ThinkGA} \cite{liu2022scinet} attempt to learn the temporal dependencies with the convolution layer. Transformer based models have also been widely applied in TSF due to the advantage of self-attention to capture the long-term dependencies of different time steps, wherein LogTrans \cite{2019Enhancing} uses convolutional self-attention layers with a LogSparse design to capture local information while lowering space complexity, Informer \cite{haoyietal-informer-2021} leverages KL-divergence based ProbSparse attention to expand the Transformer model, and reduce complexity to $O(L\log L)$, where $L$ is the length of the time series. These deep-learning models are all channel mix models, which treat different channels of a multivariate time series as a whole to fit and forecast. However, some studies \cite{han2023capacity} \cite{Zeng2022AreTE} \cite{nie2022time} have shown that channel mix may not be an appropriate manner in most cases, and have used channel independent manner to achieve good results in multivariate TSF problems.  
\subsection{Self-supervised learning for Time Series}
In order to better learn the representations of time series and apply them to downstream tasks, self-supervised learning (SSL) has also made great progress in the field of time series. The SSL methods can be roughly classified into three categories: generative-based, adversarial-based, and contrastive-based methods \cite{zhang2023selfsupervised}. The generative-based methods utilize an encoder and decoder to map and reconstruct input series, aiming to minimize reconstruction error. The adversarial-based methods employ a generator and discriminator, with the generator generating fake series and the discriminator distinguishing between real and fake series. On the other hand, the contrastive-based methods maximize mutual information between positive samples created through data augmentation or context sampling, using a contrastive similarity metric. 

In the contrastive-based methods, TS-TCC \cite{eldele2021time} applies weak augmentation and strong augmentation respectively to the original time series data. These transformations generate context vectors that are used to forecast future time steps in a mutually exclusive manner. Positive samples consist of the two augmented series derived from the same original series, while negative samples are from other series. The weights of the encoder that generates time series representations are optimized based on a contrastive loss function. TNC \cite{tonekaboni2021unsupervised} is designed for learning representations for non-stationary time series. It leverages the augmented Dickey-Fuller (ADF) test to ensure that positive samples are chosen from a neighborhood of similar signals. Furthermore, to address the issue of sampling bias, TNC employs Positive-Unlabeled (PU) learning, treating the negative samples as unknown samples and assigning appropriate weights to these samples during the learning process. TS2Vec \cite{yue2021ts2vec} is a method that utilizes hierarchical contrastive learning to generate representations for individual time steps in time series. It achieves this by generating various views of the data through masking operations, where randomly selected time steps are masked out. This enables TS2Vec to learn robust and informative representations that capture the temporal dynamics of the time series. CoST \cite{woo2022cost} is a contrastive learning-based method that encodes disentangled representations for periodic and trendy patterns in time series, facilitating more accurate analysis and forecasting tasks.

\section{Methods}
\textbf{Preliminary} Given a time series ${X=\{x^{t}_{1},...,x^{t}_{C}\}^{L}_{t=1}}$, where $L$ denotes the length of the series, and $C$ denotes the number of variables, and ${x^{t}_{i}}$ represents the value of the $i^{th}$ variable at the $t^{th}$ time step. The objective is to learn a representation function that maps ${x^{t}}$ to a representation ${r^{t}}$. Then the representation ${r^{t}}$ is used to forecast the future values of the same series, denoted as ${F=\{x^{t}_{1},...,x^{t}_{C}\}^{L+T}_{t=L+1}}$, where $T$ represents the number of time steps to forecast.

To rectify the problem of inadequate purity in time series representations encountered in previous contrastive learning methods, our proposed contrastive learning framework CLeaRForecast incorporates a variety of purifying methods from the perspectives of sample, feature, and architecture. These methods aim to obtain high-purity representations of time series for improved performance in downstream forecasting tasks. Specifically, our framework utilizes an efficient strategy for generating positive samples, a more effective training manner, and a streamlined backbone architecture combined with a global contrastive loss function. These components work together to ensure the acquisition of high-purity time series representations.
\subsection{Generation of positive sample pairs}
\begin{figure}[!t]
\centering
\includegraphics[width=1.05\columnwidth]{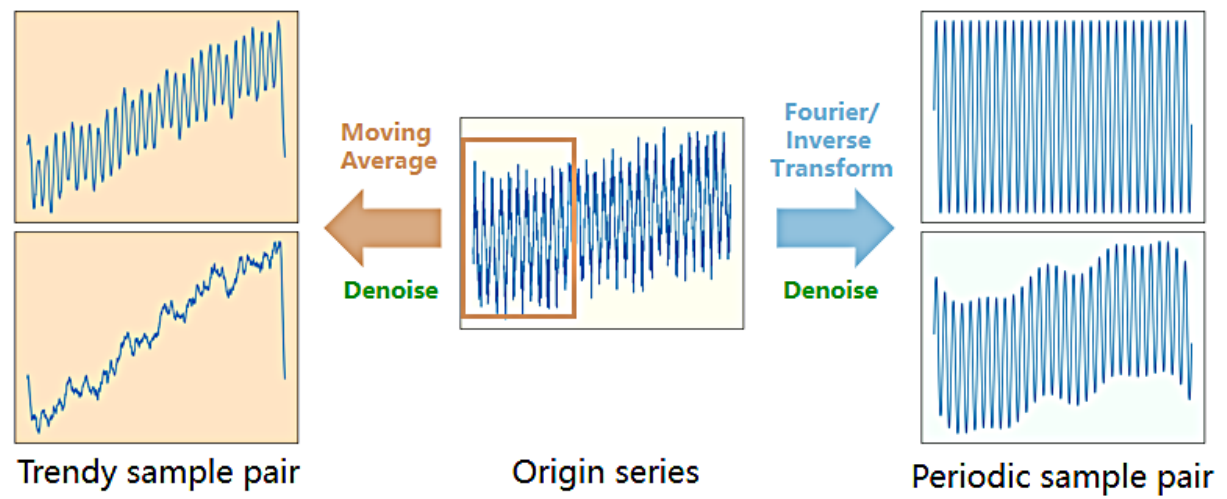} 
\caption{Generation of positive sample pairs for contrastive learning. Trendy sample pair and periodic sample pair are generated base on different trend/periodicity information of the original series.}
\label{generate samples}
\end{figure}

Unlike the methods of using the transformations of scaling, shifting and jittering \cite{eldele2021time} \cite{woo2022cost} to generate positive sample pairs of the time series, our proposed strategy to generate samples (series) sufficiently takes into account the temporal trend/periodicity information of the time series, as demonstrated in Fig. \ref{generate samples}. This strategy can eliminate noise in the time series by effectively extracting trend/periodicity features. \\ 
In terms of periodic sample generation, we employ the Fourier transform and its inverse transform operations. The Fourier transform is a common transform method that converts time-domain information into the frequency domain, with the frequency components with a high proportion of the transformed frequency domain predominantly preserving the periodicity information of the time series \cite{sneddon1995fourier}. We select different frequency components with a high proportion and subsequently invert them back to the time domain using the Inverse Fourier transform to generate the positive periodic samples for contrastive learning. In this process, the noise of time series (frequency components with a low proportion) is filtered. The process of generating a periodic sample can be defined as Eq. \ref{fft}.

\begin{eqnarray}    \label{fft}
&& A=Amp(FFT(X)),   \nonumber \\
&& k = RandInt(k1,k2), \nonumber \\
&&\{f_{1},\cdots,f_{k}\} = argTop(A,k),  \nonumber    \\
&& X_{k} = IFFT(\{f_{1},\cdots,f_{k}\}). \nonumber \\
\end{eqnarray}
where $FFT(\cdot)$ represents the Fourier transform, and $Amp(\cdot)$ denotes the calculation of amplitude values. $k$ is the number of top frequencies to be selected ($k$ is different for two positive samples generated from the same series to have similar but different periodic information), and $k1$ and $k2$ are the lower and upper bounds of random integers, which are hyperparameters. $IFFT(\cdot)$ represents the Inverse Fourier transform.

In terms of trendy sample generation, we employ sliding windows with different sizes to perform moving average operations on the original series, thereby extracting trend information of varying degrees to get the positive trendy samples for contrastive learning. Moving average is also a common method for eliminating outliers (noise) in series. The process of generating a trendy sample can be defined as Eq. \ref{moving avg}. 

\begin{eqnarray}    \label{moving avg}
&& t = RandInt(t1,t2), \nonumber \\
&& X_{t} = AvgPool(2*t+1)(X). \nonumber \\
\end{eqnarray}
where $AvgPool(\cdot)$ represents the operation of moving average, and $2*t+1$ is the size of sliding window to be selected ($t$ is also different for two positive samples generated from the same series), and $t1$ and $t2$ are the lower and upper bounds of random integers, which are also hyperparameters. The generated periodic sample and trendy sample are concatenated together to get the final positive sample. 

\subsection{Channel independent}
\begin{figure}[!t]
\centering
\includegraphics[scale=0.52]{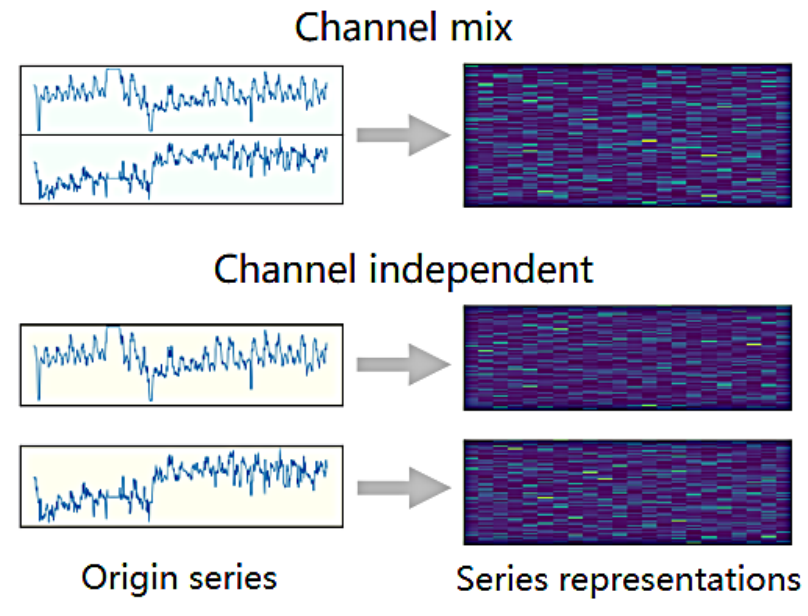} 
\caption{Channel mix manner and channel independent manner. Channel independent manner can avoid noise caused by other unrelated variables.}
\label{ind}
\end{figure}

Previous methods for representing multivariate time series data were typically channel-mix based methods. These methods assume interdependencies among different variables (channels) in the multivariate time series and treat different variables at the same time step as a single entity for learning. However, the interdependencies between different variables may not be strong. For example, different loads in the ETT datasets \cite{haoyietal-informer-2021} may not have direct influence on each other, or in Electricity data \cite{ecldata}, the electricity consumption of different customers may be very different. Therefore, treating them as a whole for learning may not be appropriate, and other unrelated variables of the series may introduce noise to the current variable. Opposite to the channel mix manner, channel independent manner treats each variable in a multivariate time series as a separate series to learn their representations, as shown in Fig. \ref{ind}. Channel independent manner allows a time series representing backbone to learn features more comprehensively and avoid noise caused by other unrelated variables for a specific series, and provides more training samples for the backbone. Many studies \cite{Zeng2022AreTE} \cite{nie2022time} have shown that channel independent methods can achieve better performance compared to channel mix methods. 

\subsection{Model Architecture}
\begin{figure*}[!htbp]
\centering
\includegraphics[scale=0.36]{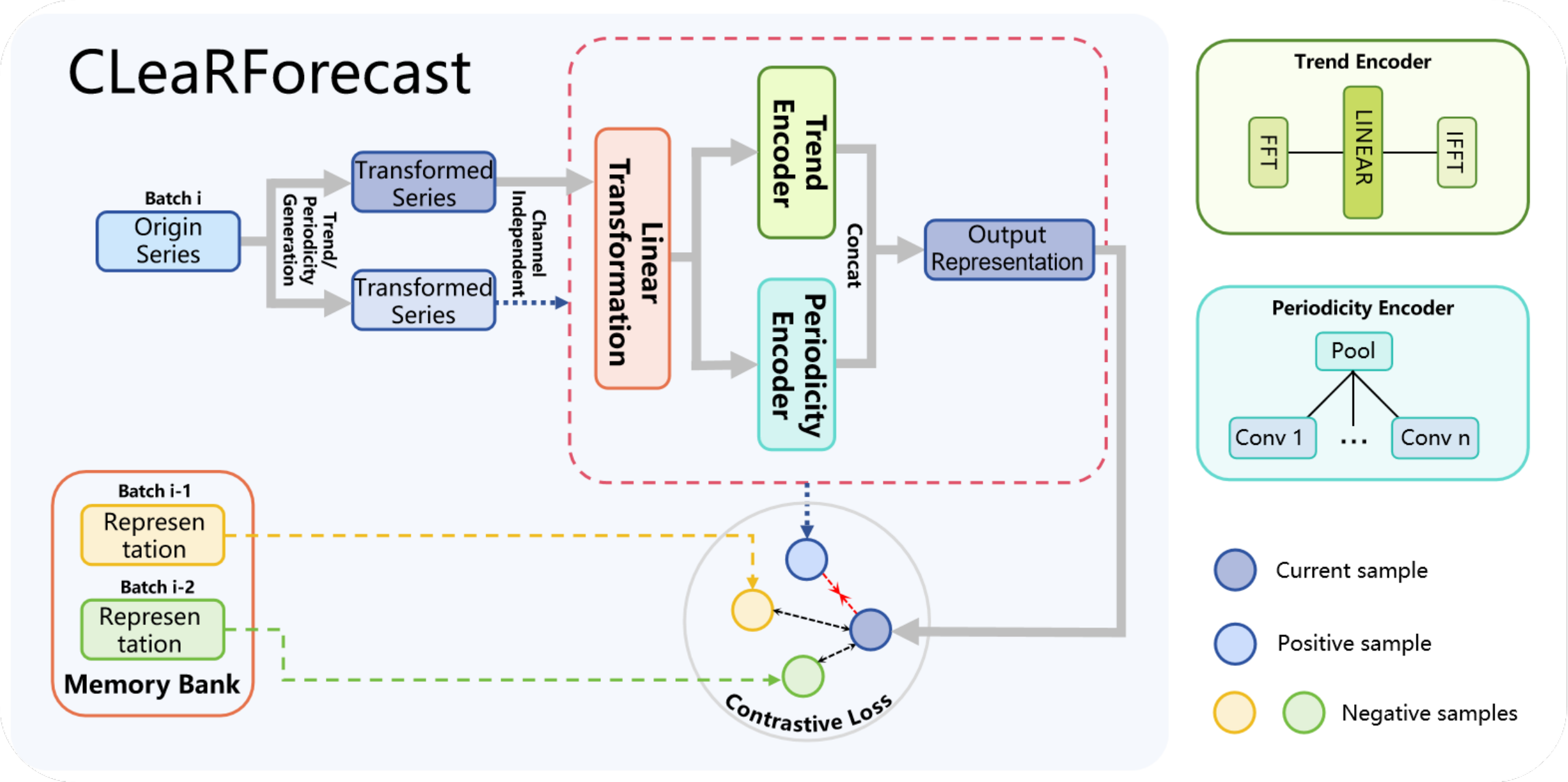} 
\caption{The architecture of CLeaRForecast. The positive sample is generated through our comparative learning sample generation strategy and linearly transformed. The periodic component and the trendy component are fed into a periodicity encoder and a trend encoder respectively to learn their representations, and the two representations are concatenated to obtain the global representation of the original series. Once the representations for both positive samples are obtained, the contrastive loss can be computed. No redundant feature extractors or contrastive loss functions to avoid additional noise.}
\label{architecture}
\end{figure*}

Fig. \ref{architecture} shows the architecture of CLeaRForecast. The transformed series (positive sample) is first generated from the original series through our contrastive learning sample generation strategy, then the series is mapped into a higher-dimensional space through a linear layer. The periodic component of the series is then fed into a periodicity encoder to learn its periodic representation, while the trend component is fed into a trend encoder to learn its trendy representation. The two representations are concatenated to obtain the global representation of the transformed series. The trend encoder consists of 1D convolutions utilizing various kernel sizes, designed to capture diverse trend information within the series. Subsequently, an average pooling layer is applied to obtain the ultimate representation of the trends. On the other hand, the periodicity encoder begins by transforming the periodic component of the series into the frequency domain using a Fourier transform layer. This transformed representation is then passed through a linear layer before being inversely transformed using the Inverse Fourier transform layer, resulting in the final periodic representation. Once the representations for both positive samples are obtained, the contrastive loss can be computed, and the deep-learning backbone (including the linear layer, trend encoder and the periodic encoder) is updated based on the loss to obtain better representations for the sample pairs.

To learn distinctive representations, we utilize a contrastive loss adapted from MoCo \cite{he2020momentum}, which is also adopted in TS2Vec and CoST. The detailed description of MoCo algorithm is shown in Appendix \ref{app:moco}. The variant of MoCo in time series incorporates a backbone and a momentum backbone for obtaining representations of positive pairs, and uses a dynamic dictionary with a queue (memory bank) for acquiring negative pairs. In the framework, series representations from the previous batches are selected as the negative samples. With $N$ samples and $K$ negative samples, the contrastive loss is expressed as:
\begin{eqnarray}    \label{moco loss}
L = \sum_{i=1}^N -\log \frac{\exp (q_i \cdot k_i / \tau)}{\sum_{j=0}^K \exp (q_i \cdot k_j / \tau)}
\end{eqnarray}

where $\tau$ is a temperature hyper-parameter, $q_i$ and $k_i$ are positive pairs, which are different representations obtained by applying different transformations to the same sample, and $q_i$ and $k_j, j \ne i$ are negative pairs. We set $K$ to twice $N$ in the experiments.

The architecture of the model is streamlined, avoiding the introduction of noise to time series representation caused by redundant feature extractors or contrastive loss functions.


\section{Experiments}
\subsection{Data and Experiment Setting}
The performance of CLeaRForecast is evaluated using five real-world datasets, including Electricity \cite{ecldata}, Weather \cite{weatherdata}, and three ETT datasets \cite{haoyietal-informer-2021}. The detatiled descriptions and statistics of the datasets can be found in Appendix \ref{app:describe}.

Five different prediction lengths are adopted for evaluation, which include 24, 48, 96, 288 and 672 for ETTm1, and 24, 48, 168, 336, and 720 for all other datasets. In line with the evaluation methodology employed in a previous research \cite{haoyietal-informer-2021}, we calculate the mean squared error (MSE) and mean absolute error (MAE) on z-score normalized data, which enables the assessment of various variables on a consistent scale.

We use two types of typical models as baselines. 1) Representation learning models: CoST \cite{woo2022cost}, TS2Vec \cite{yue2021ts2vec}, TNC \cite{tonekaboni2021unsupervised}, and a time series adaptation of MoCo \cite{he2020momentum}, these models learn representations of time series through contrastive learning and subsequently utilize ridge regressor \cite{hoerl1970ridge} to forecast future time steps based on these representations. 2) End-to-end forecasting models: Informer \cite{haoyietal-informer-2021}, LogTrans \cite{2019Enhancing}, and TCN \cite{Oord2016WaveNetAG}. The details on these baselines are shown in Appendix \ref{app:detail}.

In the experiments, we set the dimensionality of time series representation to 320. During the process of generating samples, the upper and lower bounds hyperparameters $k1$ and $k2$ for frequency selection are set to 5 and 20 respectively, and the upper and lower bounds hyperparameters $t1$ and $t2$ related to sliding window size are set to 1 and 5 respectively. We use the SGD optimizer with a learning rate between 1e-4 to 1e-3. The batch size is 256, and the default training epoch is set to 100. All experiments are run on a single GPU NVIDIA GeForce RTX 4090 with a memory of 24GB or NVIDIA Tesla A100 (only for the dataset Electricity) with a memory of 40GB. 
\subsection{Results}
The multivariate series forecasting results are shown in Table \ref{tab:main-multi}. The baseline models' results are adapted from CoST \cite{woo2022cost}. CLeaRForecast achieves the best performance on 3 ETT datasets and Electricity, and when forecasting the future for a long time, its performance is also relatively stable. All results are the average of five different random seeds test results.
\begin{table*}[!htbp]
  \centering
  \caption{Multivariate series forecasting results. Best results are highlighted in bold. The results of 5 different prediction lengths of different models are listed in the table. We also calculate the number of optimal values obtained by different models.}
  \resizebox{\textwidth}{!}{
    \input{tables/main_multi}
    }
  \label{tab:main-multi}%
\end{table*}%

The univariate series forecasting results are shown in Table \ref{tab:main-uni} of Appendix \ref{app:univariate}. CLeaRForecast's performance in univariate forecasting is also very competitive.


\subsection{Ablation Studies}
In this section, we perform ablation experiments to show that our sample generation strategy, training manner, backbone structure, and contrastive learning strategy are effective in purifying the time series representation and yielding improvements in downstream forecasting tasks. 

\textbf{Positive sample generation strategy} 
The sample generation strategy of CLeaRForecast considers the periodicity and trend information of the time series, which is more effective than the common transformations of scaling, shifting and jittering in disentangling and denoising time series. The results of replacing CLeaRForecast's sample generation strategy with the previous common transformations are shown in the second column of Table \ref{tab:abla1}.

\textbf{Channel independent vs channel mix} 
In order to demonstrate the superiority of the channel independent manner in avoiding noise caused by other variables when learning time series representations and its applicability to downstream forecasting tasks, as compared to the channel mix manner, we conducted ablation experiments using the channel mix manner. The corresponding results are presented in the third column of Table \ref{tab:abla1}.

\textbf{Trend encoder and periodicity encoder} 
Previous research \cite{woo2022cost} has shown that both trend and periodic representations of time series are important for downstream TSF tasks, and we have further validated this point. The ablation results of removing the trend encoder or periodicity encoder are shown in the fourth and fifth column of Table \ref{tab:abla1}, respectively.

\textbf{Feature extractor of dilated convolutions} 
Previous studies \cite{yue2021ts2vec} \cite{woo2022cost} have used dilated convolutions to extract features from time series to obtain a better representation of the time series. Nevertheless, our findings indicate that a more effective approach is to directly input the time series data into the trend encoder and periodicity encoder after a linear transformation, as the utilization of dilated convolutions in the backbone poses a potential risk of redundant learning of trend and periodic patterns and introducing additional noise. The results of incorporating an extra dilated convolutions for feature extraction are presented in the sixth column of Table \ref{tab:abla1}.

\textbf{Two losses or one} 
CLeaRForecast utilizes a global contrastive loss function (based on the MoCo contrastive loss) to concurrently learn both trendy and periodic representations. However, some other studies employ extra instance-wise contrastive loss functions to learn representations in other perspectives, which may result in biased learning of different patterns. The ablation results obtained using two loss functions (the MoCo-based contrastive loss function for the trendy representation, and the instance-wise contrastive loss adopted in CoST for the periodic representation) are provided in the seventh column of Table \ref{tab:abla1}.

\begin{table*}[!htbp]
  \centering
  \caption{Ablation experiments of time series representation learning. The results of replacing the sample generation strategy with previous common transformations to samples, using channel mix manner, removing the trend encoder or periodicity encoder, adding an extra feature extractor of dilated convolutions, and using two losses in place of one are listed in the table. Bold represents the best result.}
  \resizebox{\textwidth}{!}{
    \input{tables/ablation.tex}
    }
  \label{tab:abla1}%
\end{table*}%

\textbf{Ablation study of downstream forecasting tasks}
We perform an ablation study on the downstream forecasting tasks, as outlined in Table \ref{tab:abla2}. This ablation study involves substituting the original data from the corresponding time step for the high-purity learned time series representations. Subsequently, a decrease in forecasting performance is evident. Moreover, we explore alternative models such as the linear regressor \cite{su2012linear} and RandomForest regressor \cite{segal2004machine}, replacing the ridge regressor. These substitutions also lead to a moderate decline in the effectiveness of forecasting.

\begin{table}[!htbp]
  \centering
  \caption{Ablation experiments of downstream forecasting tasks. The results of using the origin series, replacing ridge regressor with linear regressor or RandomForest regressor are listed in the table. Bold represents the best result.}
  \resizebox{\columnwidth}{!}{
    \input{tables/forecast.tex}
    }
  \label{tab:abla2}%
\end{table}%
\subsection{Case study}
\begin{figure}[!t]
\centering
\includegraphics[width=1\columnwidth]{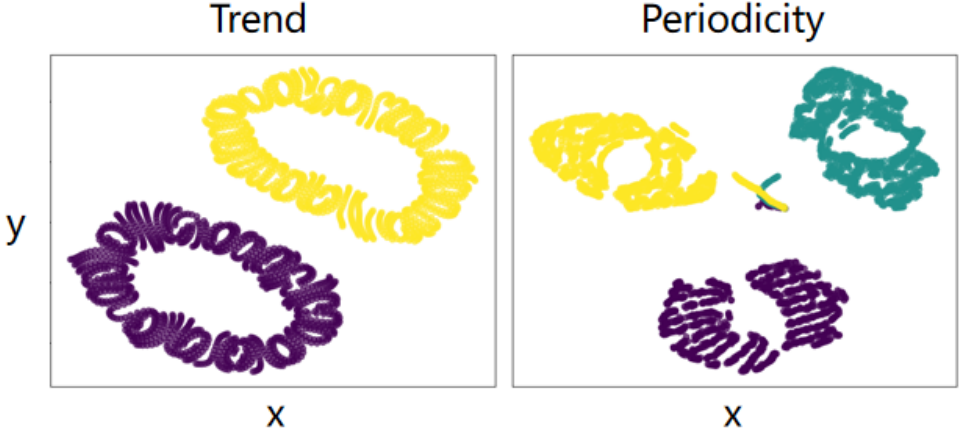} 
\caption{T-SNE visualization of learned representations from CLeaRForecast. Left sub-figure is generated by visualizing representations after selecting a single periodicity, and right sub-figure is generated by visualizing representations after selecting a single trend. Different colors represent the distinct trends or periodicities.}
\label{tsne}
\end{figure}

We construct a synthetic dataset with two distinct trend features and three different periodicity features. We employ CLeaRForecast to extract feature representations of the time series and visualize them using the T-SNE method \cite{van2008visualizing}, as shown in Fig. \ref{tsne}. It can be observed that the learned representations by CLeaRForecast successfully discriminate the different data distribution characteristics. The details of the synthetic dataset are as follows: 

\textbf{Trend features}: We construct time series data with two distinct trend features, represented by the formulas $g(t) = {\beta}_0 - ({\beta}_1 * t / {\beta}_2) + {\epsilon}_t, {\epsilon}_t \in \mathcal{N}(0, 0.3)$, where the parameters $({\beta}_0=2, {\beta}_1=1.5, {\beta}_2=500)$ and $({\beta}_0=-2, {\beta}_1=-1.5, {\beta}_2=500)$ are chosen accordingly.

\textbf{Periodicity features}: We create three different periodicity properties $p(t)$ by utilizing three sets of frequencies, phases, and amplitudes. The hyperparameters for each set are specified as $[(20, 0.0, 3), (50, 0.5, 3), (100, 1.0, 3)]$.

To generate the final synthetic dataset, we combine the trend data with the periodicity data through addition. This results in six time series sequences obtained by pairwise combinations. The length of each time series is set to 1,000.

The aforementioned demonstration highlights the capability of CLeaRForecast in disentangling periodicity and trend by effectively learning their high-purity distinct representations. However, CLeaRForecast exhibits mediocre performance when applied to time series that lack clear trend/periodicity, such as certain univariate series of the Weather data. In particular, some univariate series of the Weather data are stationary series with abrupt changes, posing challenges for our contrastive learning methods to acquire robust representations of trendy/periodic patterns, which may not even exist in such time series data. Consequently, the application of these representations to downstream forecasting tasks proves to be less effective.

\section{Conclusion}
This study introduces CLeaRForecast, a contrastive learning framework designed to learn high-purity representations of time series and effectively apply them to downstream forecasting tasks. The framework incorporates periodic and trendy representations of time series while employing purifying methods to denoise the representations, which is often overlooked in previous contrastive learning frameworks. Specifically, the proposed framework addresses several challenges of denoising by: 1) introducing a novel strategy for generating positive sample pairs that capture trend/periodicity effectively; 2) adopting a training manner of channel independent that treats each variable independently to alleviate noise induced by unrelated variables in multivariate time series; and 3) utilizing a streamlined deep-learning backbone and a global contrastive loss function to prevent redundant or biased learning of trend/periodicity, thereby preventing the introduction of noise. Experimental results demonstrate significant performance improvements when applying these high-purity and noise-free time series representations to downstream forecasting tasks. 

 \clearpage
\bibliography{aaai24}

\clearpage
\appendix
\begin{center}
\huge Supplemental Materials for  
\par
~\\
\Large CLeaRForecast: Contrastive Learning of High-Purity Representations for Time Series Forecasting
\end{center}
\section{MoCo (Momentum Contrast) Algorithm} \label{app:moco}
The MoCo \cite{he2020momentum} algorithm is a self-supervised learning framework designed to learn representations from unlabeled data. It aims to leverage the contrastive learning to effectively capture the inherent structure and semantics of data. 

MoCo consists of two encoder networks: an online encoder network and a momentum encoder network. The online encoder network is responsible for generating query representations, while the momentum encoder network serves as a dictionary or memory bank to produce key representations.

The goal of MoCo is to maximize the similarity between positive pairs while minimizing the similarity between negative pairs. This is achieved by using a contrastive loss function, such as the InfoNCE loss, which encourages the positive pairs to have higher similarity scores compared to the negative pairs. To ensure the stability and consistency of the learned representations, MoCo adopts a momentum update strategy, which involves updating the parameters of the momentum encoder network by gradually incorporating the updated parameters from the online encoder network.

MoCo has demonstrated superior performance in various computer vision tasks, including image classification, object detection, and semantic segmentation. Its effectiveness lies in its ability to learn efficient representations in a self-supervised manner, without requiring manual annotations. As the influence of MoCo continues to expand, it is playing an increasingly significant role in other fields as well.  

\section{Data Description} \label{app:describe}
The experimental data encompasses five widely-used datasets, each characterized by distinct attributes:

ETT Dataset: The ETT datasets comprises two sets of hourly-level data (ETTh1 and ETTh2) and one set of 15-minute-level data (ETTm1). These datasets consist of oil and load features of electricity transformers recorded from July 2016 to July 2018. Each dataset encompasses seven variables, with ETTh containing 17,420 timesteps and ETTm containing 69,680 timesteps. For univariate forecasting, only the "oil temperature" variable is employed for training and forecasting. In multivariate forecasting, all variables are utilized.

Electricity Dataset: This dataset comprises hourly electricity consumption (Kwh) data from 321 clients spanning the years 2012 to 2014. With 321 variables, each variable contains 26,304 timesteps. For univariate forecasting, only the "MT\_001" variable is considered for training and forecasting. In multivariate forecasting, all variables are used.

Weather Dataset: This dataset encompasses local climatological data for nearly 1,600 areas in the United States, collected over a span of four years at 10-minute intervals. Each time step captures 11 weather variables, with the target variable being "WetBulbCelsius". For univariate forecasting, only the "WetBulbCelsius" variable is taken into account, while in multivariate forecasting, all variables are utilized.

\section{Details on Baselines} \label{app:detail}
The descriptions and implementations of the seven baselines are provided below:

CoST \cite{woo2022cost}: CoST encodes disentangled representations for periodic and trendy patterns in time series, and the final representation for forecasting is obtained by concatenating the two representations. Their code is available at 
https://github.com/salesforce/CoST.

TS2Vec \cite{yue2021ts2vec}: TS2Vec adopts a hierarchical contrastive learning framework to learn a universal representation for time series. It applies timestep masks as data augmentation and temporal convolutions to encode latent representations. The available code: https://github.com/yuezhihan/ts2vec.

TNC \cite{tonekaboni2021unsupervised}: TNC presents a self-supervised framework for acquiring generalizable representations for non-stationary time series, ensuring distinguishability between neighborhood-based latent representations and representations outside the neighborhood. Their code is at 
https://github.com/sanatonek/TNC\_representation\_learning. 

MoCo \cite{he2020momentum}: MoCo (Momentum Contrast) is a well-known self-supervised learning baseline widely used in computer vision. The time series version of MoCo is adapted utilizing the publicly available source code: https://github.com/facebookresearch/moco.

Informer \cite{haoyietal-informer-2021}: Informer is an efficient transformer architecture which designed for long sequence time series forecasting. Their code is available at https://github.com/zhouhaoyi/Informer2020.

LogTrans \cite{2019Enhancing}: LogTrans employs convolutional self-attention layers with a LogSparse design to capture local information in time series while reducing complexity. The available code can be found at: https://github.com/AIStream-Peelout/flow-forecast. 

TCN \cite{Oord2016WaveNetAG}: TCN proposes dilated convolutions for time series, and their publicly available source code can be found at https://github.com/locuslab/TCN.

\section{Univariate Forecasting Results of CLeaRForecast} \label{app:univariate}
The univariate series forecasting results of different models are shown in Table \ref{tab:main-uni}. CLeaRForecast can also achieve the best results in the majority of experimental settings. All results are the average of five different random seeds test results.
\begin{table*}[!t]
  \centering
  \caption{Univariate forecasting results. Best results are highlighted in bold. The results of 5 different prediction lengths of different models are listed in the table. We also calculate the number of optimal values obtained by different models.}
  \resizebox{\textwidth}{!}{
    \input{tables/univariate_forecasting_results}
    }
  \label{tab:main-uni}%
\end{table*}%


\end{document}

%% file: tables/main_multi.tex
\begin{tabular}{cccccccccccccccccc}
\toprule
\multicolumn{1}{c}{\multirow{2}[4]{*}{Methods}} & \multicolumn{12}{c}{Representation Learning}                   & \multicolumn{4}{c}{End-to-end Forecasting} \\
\cmidrule{3-18}\multicolumn{2}{c}{} & \multicolumn{2}{c}{\textbf{CLeaRForecast}} & \multicolumn{2}{c}{CoST} & \multicolumn{2}{c}{TS2Vec} & \multicolumn{2}{c}{TNC} & \multicolumn{2}{c}{MoCo} & \multicolumn{2}{c}{Informer} & \multicolumn{2}{c}{LogTrans} & \multicolumn{2}{c}{TCN} \\
\midrule
\multicolumn{1}{c}{Metrics} & \multicolumn{1}{c}{} & MSE   & MAE   & MSE   & MAE   & MSE   & MAE   & MSE   & MAE   & MSE   & MAE   & MSE   & MAE   & MSE   & MAE & MSE   & MAE \\
\midrule
\multicolumn{1}{c|}{\multirow{5}[2]{*}{ETTh1}} & \multicolumn{1}{c|}{24} & \textbf{0.305} & \textbf{0.353}& 0.386 & 0.429 & 0.590 & 0.531 & 0.708 & 0.592 & 0.623 & \multicolumn{1}{c|}{0.555} & 0.577 & 0.549 & 0.686 & 0.604 & \multicolumn{1}{r}{0.583} & \multicolumn{1}{r}{0.547} \\
\multicolumn{1}{c|}{} & \multicolumn{1}{c|}{48} & \textbf{0.336} & \textbf{0.370} & 0.437 & 0.464 & 0.624 & 0.555 & 0.749 & 0.619 & 0.669 & \multicolumn{1}{c|}{0.586} & 0.685 & 0.625 & 0.766 & 0.757 & \multicolumn{1}{r}{0.670} & \multicolumn{1}{r}{0.606} \\
\multicolumn{1}{c|}{} & \multicolumn{1}{c|}{168} & \textbf{0.411} & \textbf{0.413} & 0.643 & 0.582 & 0.762 & 0.639 & 0.884 & 0.699 & 0.820 & \multicolumn{1}{c|}{0.674} & 0.931 & 0.752 & 1.002 & 0.846 & \multicolumn{1}{r}{0.811} & \multicolumn{1}{r}{0.680} \\
\multicolumn{1}{c|}{} & \multicolumn{1}{c|}{336} & \textbf{0.451} & \textbf{0.439} & 0.812 & 0.679 & 0.931 & 0.728 & 1.020 & 0.768 & 0.981 & \multicolumn{1}{c|}{0.755} & 1.128 & 0.873 & 1.362 & 0.952 & \multicolumn{1}{r}{1.132} & \multicolumn{1}{r}{0.815} \\
\multicolumn{1}{c|}{} & \multicolumn{1}{c|}{720} & \textbf{0.476} & \textbf{0.488} & 0.970 & 0.771 & 1.063 & 0.799 & 1.157 & 0.830 & 1.138 & \multicolumn{1}{c|}{0.831} & 1.215 & 0.896 & 1.397 & 1.291 & \multicolumn{1}{r}{1.165} & \multicolumn{1}{r}{0.813} \\
\midrule
\multicolumn{1}{c|}{\multirow{5}[2]{*}{ETTh2}} & \multicolumn{1}{c|}{24} & \textbf{0.171} & \textbf{0.268} & 0.447 & 0.502 & 0.423 & 0.489 & 0.612 & 0.595 & 0.444 & \multicolumn{1}{c|}{0.495} & 0.720 & 0.665 & 0.828 & 0.750 & \multicolumn{1}{r}{0.935} & \multicolumn{1}{r}{0.754} \\
\multicolumn{1}{c|}{} & \multicolumn{1}{c|}{48} & \textbf{0.232} & \textbf{0.314} & 0.699 & 0.637 & 0.619 & 0.605 & 0.840 & 0.716 & 0.613 & \multicolumn{1}{c|}{0.595} & 1.457 & 1.001 & 1.806 & 1.034 & \multicolumn{1}{r}{1.300} & \multicolumn{1}{r}{0.911} \\
\multicolumn{1}{c|}{} & \multicolumn{1}{c|}{168} & \textbf{0.401} & \textbf{0.429} & 1.549 & 0.982 & 1.845 & 1.074 & 2.359 & 1.213 & 1.791 & \multicolumn{1}{c|}{1.034} & 3.489 & 1.515 & 4.070 & 1.681 & \multicolumn{1}{r}{4.017} & \multicolumn{1}{r}{1.579} \\
\multicolumn{1}{c|}{} & \multicolumn{1}{c|}{336} & \textbf{0.514} & \textbf{0.498} & 1.749 & 1.042 & 2.194 & 1.197 & 2.782 & 1.349 & 2.241 & \multicolumn{1}{c|}{1.186} & 2.723 & 1.340 & 3.875 & 1.763 & \multicolumn{1}{r}{3.460} & \multicolumn{1}{r}{1.456} \\
\multicolumn{1}{c|}{} & \multicolumn{1}{c|}{720} & \textbf{0.746} & \textbf{0.616} & 1.971 & 1.092 & 2.636 & 1.370 & 2.753 & 1.394 & 2.425 & \multicolumn{1}{c|}{1.292} & 3.467 & 1.473 & 3.913 & 1.552 & \multicolumn{1}{r}{3.106} & \multicolumn{1}{r}{1.381} \\
\midrule
\multicolumn{1}{c|}{\multirow{5}[2]{*}{ETTm1}} & \multicolumn{1}{c|}{24} & \textbf{0.217} & \textbf{0.287} & 0.246 & 0.329 & 0.453 & 0.444 & 0.522 & 0.472 & 0.458 & \multicolumn{1}{c|}{0.444} & 0.323 & 0.369 & 0.419 & 0.412 & \multicolumn{1}{r}{0.363} & \multicolumn{1}{r}{0.397} \\
\multicolumn{1}{c|}{} & \multicolumn{1}{c|}{48} & \textbf{0.278} & \textbf{0.327} & 0.331 & 0.386 & 0.592 & 0.521 & 0.695 & 0.567 & 0.594 & \multicolumn{1}{c|}{0.528} & 0.494 & 0.503 & 0.507 & 0.583 & \multicolumn{1}{r}{0.542} & \multicolumn{1}{r}{0.508} \\
\multicolumn{1}{c|}{} & \multicolumn{1}{c|}{96} & \textbf{0.306} & \textbf{0.345} & 0.378 & 0.419 & 0.635 & 0.554 & 0.731 & 0.595 & 0.621 & \multicolumn{1}{c|}{0.553} & 0.678 & 0.614 & 0.768 & 0.792 & \multicolumn{1}{r}{0.666} & \multicolumn{1}{r}{0.578} \\
\multicolumn{1}{c|}{} & \multicolumn{1}{c|}{288} & \textbf{0.365} & \textbf{0.382} & 0.472 & 0.486 & 0.693 & 0.597 & 0.818 & 0.649 & 0.700 & \multicolumn{1}{c|}{0.606} & 1.056 & 0.786 & 1.462 & 1.320 & \multicolumn{1}{r}{0.991} & \multicolumn{1}{r}{0.735} \\
\multicolumn{1}{c|}{} & \multicolumn{1}{c|}{672} & \textbf{0.429} & \textbf{0.421} & 0.620 & 0.574 & 0.782 & 0.653 & 0.932 & 0.712 & 0.821 & \multicolumn{1}{c|}{0.674} & 1.192 & 0.926 & 1.669 & 1.461 & \multicolumn{1}{r}{1.032} & \multicolumn{1}{r}{0.756} \\
\midrule
\multicolumn{1}{c|}{\multirow{5}[2]{*}{Electricity}} & \multicolumn{1}{c|}{24} & \textbf{0.123} & \textbf{0.223} & 0.136 & 0.242 & 0.287 & 0.375 & 0.354 & 0.423 & 0.288 & \multicolumn{1}{c|}{0.374} & 0.312 & 0.387 & 0.297 & 0.374 & \multicolumn{1}{r}{0.235} & \multicolumn{1}{r}{0.346} \\
\multicolumn{1}{c|}{} & \multicolumn{1}{c|}{48} & \textbf{0.141} & \textbf{0.240} & 0.153 & 0.258 & 0.309 & 0.391 & 0.376 & 0.438 & 0.310 & \multicolumn{1}{c|}{0.390} & 0.392 & 0.431 & 0.316 & 0.389 & \multicolumn{1}{r}{0.253} & \multicolumn{1}{r}{0.359} \\
\multicolumn{1}{c|}{} & \multicolumn{1}{c|}{168} & \textbf{0.167} & \textbf{0.262} & 0.175 & 0.275 & 0.335 & 0.410 & 0.402 & 0.456 & 0.337 & \multicolumn{1}{c|}{0.410} & 0.515 & 0.509 & 0.426 & 0.466 & \multicolumn{1}{r}{0.278} & \multicolumn{1}{r}{0.372} \\
\multicolumn{1}{c|}{} & \multicolumn{1}{c|}{336} & \textbf{0.185} & \textbf{0.281} & 0.196 & 0.296 & 0.351 & 0.422 & 0.417 & 0.466 & 0.353 & \multicolumn{1}{c|}{0.422} & 0.759 & 0.625 & 0.365 & 0.417 & \multicolumn{1}{r}{0.287} & \multicolumn{1}{r}{0.382} \\
\multicolumn{1}{c|}{} & \multicolumn{1}{c|}{720} &\textbf{0.223} & \textbf{0.315} & 0.232 & 0.327 & 0.378 & 0.440 & 0.442 & 0.483 & 0.380 & \multicolumn{1}{c|}{0.441} & 0.969 & 0.788 & 0.344 & 0.403 & \multicolumn{1}{r}{0.287} & \multicolumn{1}{r}{0.381} \\
\midrule
\multicolumn{1}{c|}{\multirow{5}[2]{*}{Weather}} & \multicolumn{1}{c|}{24} & 0.349 & 0.387 & \textbf{0.298} & \textbf{0.360} & 0.307 & 0.363 & 0.320 & 0.373 & 0.311 & \multicolumn{1}{c|}{0.365} & 0.335 & 0.381 & 0.435 & 0.477 & \multicolumn{1}{r}{0.321} & \multicolumn{1}{r}{0.367} \\
\multicolumn{1}{c|}{} & \multicolumn{1}{c|}{48} & 0.420 & 0.445 & \textbf{0.359} & \textbf{0.411} & 0.374 & 0.418 & 0.380 & 0.421 & 0.372 & \multicolumn{1}{c|}{0.416} & 0.395 & 0.459 & 0.426 & 0.495 & \multicolumn{1}{r}{0.386} & \multicolumn{1}{r}{0.423} \\
\multicolumn{1}{c|}{} & \multicolumn{1}{c|}{168} & 0.527 & 0.525 & \textbf{0.464} & \textbf{0.491} & 0.491 & 0.506 & 0.479 & 0.495 & 0.482 & \multicolumn{1}{c|}{0.499} & 0.608 & 0.567 & 0.727 & 0.671 & \multicolumn{1}{r}{0.491} & \multicolumn{1}{r}{0.501} \\
\multicolumn{1}{c|}{} & \multicolumn{1}{c|}{336} & 0.561 & 0.550 & \textbf{0.497} & 0.517 & 0.525 & 0.530 & 0.505 & 0.514 & 0.516 & \multicolumn{1}{c|}{0.523} & 0.702 & 0.620 & 0.754 & 0.670 & \multicolumn{1}{r}{0.502} & \multicolumn{1}{r}{\textbf{0.507}} \\
\multicolumn{1}{c|}{} & \multicolumn{1}{c|}{720} & 0.625 &  0.591 & 0.533 & 0.542 & 0.556 & 0.552 & 0.519 & 0.525 & 0.540 & \multicolumn{1}{c|}{0.540} & 0.831 & 0.731 & 0.885 & 0.773 & \multicolumn{1}{r}{\textbf{0.498}} & \multicolumn{1}{r}{\textbf{0.508}} \\
\midrule
\multicolumn{1}{c}{{1st Count}} & & \multicolumn{2}{c}{\textbf{40}} & \multicolumn{2}{c}{7} & \multicolumn{2}{c}{0} & \multicolumn{2}{c}{0} & \multicolumn{2}{c|}{0} & \multicolumn{2}{c}{0} & \multicolumn{2}{c}{0} & \multicolumn{2}{c}{3}\\
\bottomrule
\end{tabular}%

%% file: tables/ablation.tex
  \begin{tabular}{c|c|cc|cc|cc|cc|cc|cc|cc}
    \toprule
    \multicolumn{2}{c}{Models} &
    \multicolumn{2}{c}{\textbf{CLeaRForecast}} &
    \multicolumn{2}{c}{Common Trans} &
    \multicolumn{2}{c}{Channel mix} &
    \multicolumn{2}{c}{- Trend} &
    \multicolumn{2}{c}{- Periodicity} &
    \multicolumn{2}{c}{+ Dilated Conv} &
    \multicolumn{2}{c}{Two losses}
    \\
    \cmidrule(lr){3-4} 
    \cmidrule(lr){5-6}
    \cmidrule(lr){7-8} 
    \cmidrule(lr){9-10}
    \cmidrule(lr){11-12}
    \cmidrule(lr){13-14}
    \cmidrule(lr){15-16}
    \multicolumn{2}{c}{Metric} & MSE & MAE & MSE & MAE & MSE & MAE & MSE & MAE & MSE & MAE & MSE & MAE & MSE & MAE \\
    \toprule
    \multirow{4}{*}{\rotatebox{90}{ETTh1}} 
    & 24  & \textbf{0.305} & \textbf{0.353} & 0.314 & 0.357  & 0.344 & 0.392 & 0.316 & 0.360 & 0.309 & 0.350 & 0.314 & 0.359 & 0.311 & 0.356\\
    & 48  & \textbf{0.336} & \textbf{0.370} & 0.342 & 0.373  & 0.393 & 0.424 & 0.344 & 0.375 & 0.343 & \textbf{0.370} & 0.347 & 0.377 & 0.341 & 0.372\\
    & 168 & \textbf{0.411} & \textbf{0.413} & \textbf{0.411} & \textbf{0.413} & 0.467 & 0.447 & 0.415 & 0.414 & 0.427 & 0.422 & 0.421 & 0.418 & \textbf{0.411} & \textbf{0.413}\\
    & 336 & \textbf{0.451} & \textbf{0.439} & \textbf{0.451} & 0.440  & 0.840 & 0.697 & 0.452 & 0.440 & 0.487 & 0.459 & 0.463 & 0.446 & \textbf{0.451} & \textbf{0.439}\\
    & 720 & \textbf{0.476} & 0.488 & 0.477 & \textbf{0.486}  & 1.001 & 0.794 & 0.477 & 0.489 & 0.517 & 0.508 & 0.494 & 0.498 & 0.477 & 0.488 \\
    \midrule
    \multirow{4}{*}{\rotatebox{90}{ETTh2}} 
    & 24  & \textbf{0.171} & \textbf{0.268} & 0.173 & 0.270  & 0.331 & 0.417 & 0.173 & 0.270 & 0.185 & 0.282 & 0.173 & 0.270 & 0.172 & 0.269 \\
    & 48  & \textbf{0.232} & \textbf{0.314} & 0.236 & 0.317  & 0.546 & 0.552 & 0.233 & 0.315 & 0.255 & 0.336 & 0.241 & 0.320 & 0.233 & 0.315\\
    & 168 & \textbf{0.401} & \textbf{0.429} & 0.411 & 0.435  & 1.524 & 0.959 & 0.402 & 0.430 & 0.462 & 0.468 & 0.418 & 0.442 & 0.405 & 0.432\\
    & 336 & 0.514 & 0.498 & 0.534 & 0.509  & 1.907 & 1.091 & 0.515 & 0.499 & 0.616 & 0.551 & \textbf{0.502} & \textbf{0.495} & 0.523 & 0.503 \\
    & 720 & \textbf{0.746} & \textbf{0.616} & 0.759 & 0.624  & 2.055 & 1.132 & 0.747 & 0.617 & 0.837 & 0.662 & 0.946 & 0.678 & 0.751 & 0.619\\
    \midrule
    \multirow{4}{*}{\rotatebox{90}{ETTm1}} 
    & 96  & \textbf{0.217} & \textbf{0.287} & 0.221 & 0.292  & 0.236 & 0.318 & 0.220 & 0.291 & 0.425 & 0.407 & 0.239 & 0.308 & 0.221 & 0.293 \\
    & 192 & \textbf{0.278} & \textbf{0.327} & \textbf{0.278} & 0.329  & 0.310 & 0.366 & 0.279 & 0.329 & 0.419 & 0.410 & 0.287 & 0.338 & \textbf{0.278} & 0.329\\
    & 336 & \textbf{0.306} & \textbf{0.345} & 0.307 & 0.346  & 0.354 & 0.396 & 0.307 & 0.346 & 0.400 & 0.402 & 0.312 & 0.352 & 0.307 & 0.346\\
    & 720 & \textbf{0.365} & \textbf{0.382} & 0.367 & 0.384  & 0.456 & 0.466 & 0.366 & 0.383 & 0.461 & 0.438 & 0.370 & 0.386 & 0.367 & 0.384\\
    & 720 & \textbf{0.429} & \textbf{0.421} & 0.431 & 0.423  & 0.624 & 0.566 & \textbf{0.429} & 0.422 & 0.520 & 0.475 & 0.437 & 0.426 & 0.431 & 0.423 \\
    \bottomrule
  \end{tabular}

%% file: tables/forecast.tex
\begin{tabular}{c|c|cc|cc|cc|cc}
    \toprule
    \multicolumn{2}{c}{Models} &
    \multicolumn{2}{c}{\textbf{Ridge}} &
    \multicolumn{2}{c}{Origin data} &
    \multicolumn{2}{c}{Linear} &
    \multicolumn{2}{c}{RandomForest} 
    \\
    \cmidrule(lr){3-4} 
    \cmidrule(lr){5-6}
    \cmidrule(lr){7-8} 
    \cmidrule(lr){9-10}
    \multicolumn{2}{c}{Metric} & MSE & MAE & MSE & MAE & MSE & MAE & MSE & MAE \\ 
    \toprule
    \multirow{4}{*}{\rotatebox{90}{ETTh1}} 
    & 24  & \textbf{0.305} & \textbf{0.353} & 0.875 & 0.627  & 0.315 & 0.358 & 0.374 & 0.411 \\
    & 48 & \textbf{0.336} & \textbf{0.370} & 0.893 & 0.644  & 0.343 & 0.374 & 0.412 & 0.435 \\
    & 168 & \textbf{0.411} & \textbf{0.413} & 0.914 & 0.669  & 0.413 & 0.415 & 0.487 & 0.480 \\
    & 336 & \textbf{0.451} & \textbf{0.439} & 0.916 & 0.684 & 0.453 & 0.441 & 0.532 & 0.506 \\
    & 720 & \textbf{0.476} & \textbf{0.488} & 0.929 & 0.711 & 0.479 & 0.489  & 0.577 & 0.551 \\
    
    \bottomrule
  \end{tabular}

%% file: tables/univariate_forecasting_results.tex
\begin{tabular}{cccccccccccccccccc}
\toprule
\multicolumn{1}{c}{\multirow{2}[4]{*}{Methods}} & \multicolumn{12}{c}{Representation Learning}                   & \multicolumn{4}{c}{End-to-end Forecasting}   \\
\cmidrule{3-18}\multicolumn{2}{c}{} & \multicolumn{2}{c}{\textbf{CLeaRForecast}}& \multicolumn{2}{c}{CoST} & \multicolumn{2}{c}{TS2Vec} & \multicolumn{2}{c}{TNC} & \multicolumn{2}{c}{MoCo} & \multicolumn{2}{c}{Informer} & \multicolumn{2}{c}{LogTrans} & \multicolumn{2}{c}{TCN}\\
\midrule
\multicolumn{1}{c}{Metrics} & & MSE   & MAE   & MSE   & MAE   & MSE   & MAE   & MSE   & MAE   & MSE   & MAE   & MSE   & MAE   & MSE   & MAE   & MSE   & MAE \\
\midrule
\multicolumn{1}{c|}{\multirow{5}[2]{*}{ETTh1}} & \multicolumn{1}{c|}{24} & \textbf{0.027} & \textbf{0.124} & 0.040 & 0.152 & 0.039 & 0.151 & 0.057 & 0.184 & 0.040 & \multicolumn{1}{c|}{\textbf{0.151}} & 0.098 & 0.247 & 0.103 & 0.259 & \multicolumn{1}{r}{0.104} & \multicolumn{1}{r}{0.254}  \\
\multicolumn{1}{c|}{} & \multicolumn{1}{c|}{48} & \textbf{0.041} & \textbf{0.151} & 0.060 & 0.186 & 0.062 & 0.189 & 0.094 & 0.239 & 0.063 & \multicolumn{1}{c|}{0.191} & 0.158 & 0.319 & 0.167 & 0.328 & \multicolumn{1}{r}{0.206} & \multicolumn{1}{r}{0.366}  \\
\multicolumn{1}{c|}{} & \multicolumn{1}{c|}{168} & \textbf{0.078} & \textbf{0.201} & 0.097 & 0.236 & 0.142 & 0.291 & 0.171 & 0.329 & 0.122 & \multicolumn{1}{c|}{0.268} & 0.183 & 0.346 & 0.207 & 0.375 & \multicolumn{1}{r}{0.462} & \multicolumn{1}{r}{0.586}  \\
\multicolumn{1}{c|}{} & \multicolumn{1}{c|}{336} & \textbf{0.096} & \textbf{0.240} & 0.112 & 0.258 & 0.160 & 0.316 & 0.192 & 0.357 & 0.144 & \multicolumn{1}{c|}{0.297} & 0.222 & 0.387 & 0.230 & 0.398 & \multicolumn{1}{r}{0.422} & \multicolumn{1}{r}{0.564}  \\
\multicolumn{1}{c|}{} & \multicolumn{1}{c|}{720} & 0.167 & 0.335 & \textbf{0.148} & \textbf{0.306} & 0.179 & 0.345 & 0.235 & 0.408 & 0.183 & \multicolumn{1}{c|}{0.347} & 0.269 & 0.435 & 0.273 & 0.463 & \multicolumn{1}{r}{0.438} & \multicolumn{1}{r}{0.578}  \\
\midrule
\multicolumn{1}{c|}{\multirow{5}[2]{*}{ETTh2}} & \multicolumn{1}{c|}{24} & \textbf{0.065} & \textbf{0.188} & 0.079 & 0.207 & 0.091 & 0.230 & 0.097 & 0.238 & 0.095 & \multicolumn{1}{c|}{0.234} & 0.093 & 0.240 & 0.102 & 0.255 & \multicolumn{1}{r}{0.109} & \multicolumn{1}{r}{0.251}  \\
\multicolumn{1}{c|}{} & \multicolumn{1}{c|}{48} & \textbf{0.094} & \textbf{0.232} & 0.118 & 0.259 & 0.124 & 0.274 & 0.131 & 0.281 & 0.130 & \multicolumn{1}{c|}{0.279} & 0.155 & 0.314 & 0.169 & 0.348 & \multicolumn{1}{r}{0.147} & \multicolumn{1}{r}{0.302}  \\
\multicolumn{1}{c|}{} & \multicolumn{1}{c|}{168} & \textbf{0.168} & \textbf{0.318} & 0.189 & 0.339 & 0.198 & 0.355 & 0.197 & 0.354 & 0.204 & \multicolumn{1}{c|}{0.360} & 0.232 & 0.389 & 0.246 & 0.422 & \multicolumn{1}{r}{0.209} & \multicolumn{1}{r}{0.366}  \\
\multicolumn{1}{c|}{} & \multicolumn{1}{c|}{336} & 0.218 & 0.372 & \textbf{0.206} & \textbf{0.360} & 0.205 & 0.364 & 0.207 & 0.366 & 0.206 & \multicolumn{1}{c|}{0.364} & 0.263 & 0.417 & 0.267 & 0.437 & \multicolumn{1}{r}{0.237} & \multicolumn{1}{r}{0.391}  \\
\multicolumn{1}{c|}{} & \multicolumn{1}{c|}{720} & 0.303 & 0.450 & 0.214 & 0.371 & 0.208 & 0.371 & 0.207 & 0.370 & 0.206 & \multicolumn{1}{c|}{0.369} & 0.277 & 0.431 & 0.303 & 0.493 & \multicolumn{1}{r}{\textbf{0.200}} & \multicolumn{1}{r}{\textbf{0.367}}  \\
\midrule
\multicolumn{1}{c|}{\multirow{5}[2]{*}{ETTm1}} & \multicolumn{1}{c|}{24} & \textbf{0.010} & \textbf{0.074} & 0.015 & 0.088 & 0.016 & 0.093 & 0.019 & 0.103 & 0.015 & \multicolumn{1}{c|}{0.091} & 0.030 & 0.137 & 0.065 & 0.202 & \multicolumn{1}{r}{0.027} & \multicolumn{1}{r}{0.127}  \\
\multicolumn{1}{c|}{} & \multicolumn{1}{c|}{48} &  \textbf{0.018} & \textbf{0.099} & 0.025 & 0.117 & 0.028 & 0.126 & 0.036 & 0.142 & 0.027 & \multicolumn{1}{c|}{0.122} & 0.069 & 0.203 & 0.078 & 0.220 & \multicolumn{1}{r}{0.040} & \multicolumn{1}{r}{0.154}  \\
\multicolumn{1}{c|}{} & \multicolumn{1}{c|}{96} & \textbf{0.029} & \textbf{0.125} & 0.038 & 0.147 & 0.045 & 0.162 & 0.054 & 0.178 & 0.041 & \multicolumn{1}{c|}{0.153} & 0.194 & 0.372 & 0.199 & 0.386 & \multicolumn{1}{r}{0.097} & \multicolumn{1}{r}{0.246} \\
\multicolumn{1}{c|}{} & \multicolumn{1}{c|}{288} & \textbf{0.056} & \textbf{0.175} & 0.077 & 0.209 & 0.095 & 0.235 & 0.098 & 0.244 & 0.083 & \multicolumn{1}{c|}{0.219} & 0.401 & 0.554 & 0.411 & 0.572 & \multicolumn{1}{r}{0.305} & \multicolumn{1}{r}{0.455}  \\
\multicolumn{1}{c|}{} & \multicolumn{1}{c|}{672} & \textbf{0.084} & \textbf{0.215} & 0.113 & 0.257 & 0.142 & 0.290 & 0.136 & 0.290 & 0.122 & \multicolumn{1}{c|}{0.268} & 0.512 & 0.644 & 0.598 & 0.702 & \multicolumn{1}{r}{0.445} & \multicolumn{1}{r}{0.576}  \\
\midrule
\multicolumn{1}{c|}{\multirow{5}[2]{*}{Electricity}} & \multicolumn{1}{c|}{24} & \textbf{0.238}     & \textbf{0.248} & 0.243 & 0.264 & 0.260 & 0.288 & 0.252 & 0.278 & 0.254 & \multicolumn{1}{c|}{0.280} & 0.251 & 0.275 & 0.528 & 0.447 & \multicolumn{1}{r}{0.243} & \multicolumn{1}{r}{0.367}  \\
\multicolumn{1}{c|}{} & \multicolumn{1}{c|}{48} & \textbf{0.281} & \textbf{0.280} & 0.292 & 0.300 & 0.313 & 0.321 & 0.300 & 0.308 & 0.304 & \multicolumn{1}{c|}{0.314} & 0.346 & 0.339 & 0.409 & 0.414 & \multicolumn{1}{r}{0.283} & \multicolumn{1}{r}{0.397}  \\
\multicolumn{1}{c|}{} & \multicolumn{1}{c|}{168} & 0.390   & \textbf{0.364} & 0.405 & 0.375 & 0.429 & 0.392 & 0.412 & 0.384 & 0.416 & \multicolumn{1}{c|}{0.391} & 0.544 & 0.424 & 0.959 & 0.612 & \multicolumn{1}{r}{\textbf{0.357}} & \multicolumn{1}{r}{0.449}  \\
\multicolumn{1}{c|}{} & \multicolumn{1}{c|}{336} & 0.534     & 0.464 & 0.560 & 0.473 & 0.565 & 0.478 & 0.548 & 0.466 & 0.556 & \multicolumn{1}{c|}{0.482} & 0.713 & 0.512 & 1.079 & 0.639 & \multicolumn{1}{r}{\textbf{0.355}} & \multicolumn{1}{r}{\textbf{0.446}}  \\
\multicolumn{1}{c|}{} & \multicolumn{1}{c|}{720} & 0.842  & 0.644 & 0.889 & 0.645 & 0.863 & 0.651 & 0.859 & 0.651 & 0.858 & \multicolumn{1}{c|}{0.653} & 1.182 & 0.806 & 1.001 & 0.714 & \multicolumn{1}{r}{\textbf{0.387}} & \multicolumn{1}{r}{\textbf{0.477}}  \\
\midrule
\multicolumn{1}{c|}{\multirow{5}[2]{*}{Weather}} & \multicolumn{1}{c|}{24} & \textbf{0.096} & \textbf{0.213} & \textbf{0.096} & \textbf{0.213} & \textbf{0.096} & 0.215 & 0.102 & 0.221 & 0.097 & \multicolumn{1}{c|}{0.216} & 0.117 & 0.251 & 0.136 & 0.279 & \multicolumn{1}{r}{0.109} & \multicolumn{1}{r}{0.217}  \\
\multicolumn{1}{c|}{} & \multicolumn{1}{c|}{48} & \textbf{0.138} & \textbf{0.261} & \textbf{0.138} & 0.262 & 0.140 & 0.264 & 0.139 & 0.264 & 0.140 & \multicolumn{1}{c|}{0.264} & 0.178 & 0.318 & 0.206 & 0.356 & \multicolumn{1}{r}{0.143} & \multicolumn{1}{r}{0.269}  \\
\multicolumn{1}{c|}{} & \multicolumn{1}{c|}{168} & 0.237 & 0.354 & 0.207 & 0.334 & 0.207 & 0.335 & 0.198 & 0.328 & 0.198 & \multicolumn{1}{c|}{0.326} & 0.266 & 0.398 & 0.309 & 0.439 & \multicolumn{1}{r}{\textbf{0.188}} & \multicolumn{1}{r}{\textbf{0.319}}  \\
\multicolumn{1}{c|}{} & \multicolumn{1}{c|}{336} & 0.288 & 0.395 & 0.230 & 0.356 & 0.231 & 0.360 & 0.215 & 0.347 & 0.220 & \multicolumn{1}{c|}{0.350} & 0.297 & 0.416 & 0.359 & 0.484 & \multicolumn{1}{r}{\textbf{0.192}} & \multicolumn{1}{r}{\textbf{0.320}}  \\
\multicolumn{1}{c|}{} & \multicolumn{1}{c|}{720} & 0.386 & 0.474 & 0.242 & 0.370 & 0.233 & 0.365 & 0.219 & 0.353 & 0.224 & \multicolumn{1}{c|}{0.357} & 0.359 & 0.466 & 0.388 & 0.499 & \multicolumn{1}{r}{\textbf{0.198}} & \multicolumn{1}{r}{\textbf{0.329}}  \\
\midrule
\multicolumn{1}{c}{{1st Count}} & & \multicolumn{2}{c}{\textbf{33}} & \multicolumn{2}{c}{7} & \multicolumn{2}{c}{1} & \multicolumn{2}{c}{0} & \multicolumn{2}{c|}{0} & \multicolumn{2}{c}{0} & \multicolumn{2}{c}{0} & \multicolumn{2}{c}{13}\\
\bottomrule
\end{tabular}%